TITLE

The relationship between linguistic expression and symptoms of depression, anxiety, and suicidal thoughts: A longitudinal study of blog content

AUTHORS

O'Dea B. [1], Boonstra T.W. [1], Larsen M.E.[1]., Nguyen T. [2], Venkatesh S. [2], Christensen H. [1]

AFFILIATIONS

[1]Black Dog Institute, Faculty of Medicine, University of New South Wales, NSW, Australia

[2]Centre for Pattern Recognition and Data Analytics, Deakin University, Victoria, Australia

ABSTRACT

**Introduction:** Due to its popularity and availability, social media data may present a new way to identify individuals who are experiencing mental illness. By analysing individuals' blog content, this study aimed to investigate the associations between linguistic features and symptoms of depression, generalised anxiety, and suicidal ideation.

**Methods:** This study utilised a longitudinal study design. Individuals who blogged were invited to participate in a study in which they completed fortnightly mental health questionnaires (PHQ-9, GAD-7) for a period of 36 weeks. Linguistic features were extracted from blog data using the LIWC tool. Bivariate and multivariate analyses were performed to investigate the correlations between the linguistic features and mental health scores between subjects. We then used the multivariate regression model to predict longitudinal changes in mood within subjects.

**Results:** A total of 153 participants consented to taking part, with 38 participants completing the required number of questionnaires and blog posts during the study period. Between-subject analysis revealed that several linguistic features, including tentativeness and non-fluencies, were significantly associated with depression and anxiety symptoms, but not suicidal thoughts. Within-subject analysis showed no robust correlations between linguistic features and changes in mental health score.

**Discussion:** This study provides further support for the relationship between linguistic features within social media data and individuals' symptoms of depression and anxiety. The lack of robust within-subject correlations indicate that the relationship observed at the group level may not generalise to individual changes over time.

# INTRODUCTION

Worldwide, mental illness is a leading cause of disability and represents a major health and economic burden (1). This is due in part to the detrimental effects of mental illness on functioning, but also the low levels of mental health literacy among individuals, their inability to recognise symptoms, poor help-seeking attitudes, and lack of access to care (2-4). A fatal and tragic outcome of poor mental health is suicide, a primary cause of death for both young (5) and middle-aged people in many developed countries. There is a need to look to new ways of detecting mental illness in the population, particularly in the prodromal phase, to increase treatment outcomes, reduce severity, and prevent death (6). Social media has emerged as a potential means for doing this (7).

Defined as any internet-enabled platform that allows individuals to connect, communicate, and share content with others, social media includes social networking sites (e.g. Facebook), microblogs (e.g. Twitter), blogsites (e.g. WordPress, LiveJournal), and communities (e.g. Reddit) (8). There has been significant enthusiasm in the potential of these platforms to generate markers of poor mental health as they are used by millions of people worldwide, data is produced in natural settings and is readily available and free. It has been hypothesised that the language and expressive features within individuals' shared content may indicate their mental state (9). This is based on psycho-linguistic theory which postulates that the words and features used in everyday language can reveal individuals' thoughts, emotions, and motivations (10-12). Several promising findings have emerged.

On Twitter, De Choudhury et al (13) was able to discriminate current depression among users by their increased use of first-person pronouns and fewer references to third persons. Statistical modelling was most accurate when only linguistic features were used. Using cross-validation methods, Reece et al (14) found that depression among Twitter users was predicted by differences in word count, references to ingestion, sadness, swear words, article words, and positive emotion. Among a sample of Japanese Twitter uses, Tsugawa et al (15) found depressed users had significantly higher ratios of negative emotion words. When examining Twitter posts that included depression terms, Wilson et al (16) found these to be characterised by higher character counts, fewer pronouns, less positive emotion, greater negative emotion, greater expressions of sadness, fewer references to time and fewer references to past and present tense. When comparing posts made in online depression forums with those in breast cancer forums, Ramirez-Esparza et al (17) found the depression posts were characterised by greater first-person referencing, less positive emotion, and greater negative emotion when compared to tweets related to breast cancer. On Facebook, Seabrook et al (18) found depression was characterised by differences in the proportion of negative emotions whereas on Twitter, depression was associated with less dispersion of negative emotion across posts. Also on Facebook, Eichstaedt et al (19) found depression to be marked by the linguistic features of increased first-person pronoun use, greater negative emotion, increased perceptual processes for feeling (e.g. feels, touch), greater references to sadness (e.g. crying, grief), and greater discrepancies (e.g. should, would, could). Our team (20) discriminated higher risk in suicide-related Twitter posts by greater self-references, anger, and a focus on the present (9). Importantly, the findings are consistent with a recent meta-analysis which firmly established first person singular pronoun use as a linguistic marker of depression (21). Together, these studies provide strong support for the potential of analysing social media content for the automatic detection of mental health problems.

However, if we wish to use social media to screen and monitor the mental health of individuals, the field must first overcome major methodological challenges. Firstly, not all prior studies have correlated the various linguistic features against a validated psychometric measure of mental health. A review of recent papers (22) found that of the 12 included studies, only five used mental health questionnaires (13-15, 23, 24). The remaining relied on self-declared diagnoses (e.g. affirmative statements of mental health diagnoses in social media posts), membership association (e.g. belonging to a certain online community or forum), or annotation of content (e.g. presence of key words or phrases). As such, it is not clear whether all past findings are consistent with diagnostic criteria. Secondly, most of the studies in this area have focussed only on depression. Although this is warranted, due to the prevalence, little attention has been paid to other mental health symptoms such as anxiety, which is highly correlated with depression (25). Detecting symptoms of anxiety may be an effective way to identify those who have depression or who are at risk of developing it (26). Thirdly, current knowledge is mostly based on cross-sectional studies, with temporal patterns in depressive symptomatology and suicidal thinking unaccounted for (27). Cross-sectional studies identify markers at the group level and hence these markers may not apply to changes within individuals (28-30). Indeed, the field of personalised medicine argues that individuals may have unique markers of mental ill-health (31). It is unclear whether the markers derived from social media data can be used to infer data about the mental health state of individual users.

The current study aims to overcome some of the above limitations by collecting validated mental health data in a longitudinal study of individuals who blog. Using the text content extracted from their blogsite, alongside fortnightly mental health assessments using standard validated questionnaires, this study examines the relationship between different linguistic features and symptoms of depression, anxiety, and suicidal ideation. As found in past studies, it is hypothesised that higher mental health scores will be associated with increased references to oneself, increased expressions of negative emotion, and reduced references to third-persons at group level. This study will also test whether the group-level correlations can be used to make predictions about intra-individual changes in mental health scores over time. It is hoped that these outcomes will help us to establish prediction models which can be used to monitor blog sites automatically, and in real time, for mental health risk.

**METHOD**

**Study design.** A 36-week longitudinal cohort study approved by UNSW Human Research Ethics Committee (#14086). Recruitment took place between July 2014 and October 2016. Using a series of online adverts published on various social media channels, individuals who blogged were invited to visit the study website where they were provided with participant information and consent details. There were no exclusion criteria, although participants had to provide the URL of their blog site and self-identify as a mental health blogger. Once consent was given, participants were asked to complete an online self-report mental health assessment at baseline, and then every fortnight via email for a period of 36 weeks (total of 18 assessments). Support contacts were provided to the entire sample and there were no restrictions on help-seeking behaviour throughout the study. Each participant received $20AUD in Amazon Webstore credit for their participation.

**Measures**. *Demographics* were assessed using questions on age, gender, prior diagnosis of depression or anxiety from a medical practitioner, and medication use for depression and anxiety. *Overall health rating:* Participants were asked to rate their health as very bad, bad, moderate, good, very good. *Depressive symptoms* were assessed using the self-report Patient Health Questionnaire (PHQ-9) (32). This nine-item questionnaire assessed the presence of depressive symptoms in the past two weeks. Individuals were asked to rate the frequency of depressive symptoms using a four-point Likert scale ranging from "none of the time" to "every day or almost every day". A total score is then calculated which can be classified as "nil-minimal" (0-4), "mild" (5-9), "moderate" (10-14), "moderately severe" (15-19) or "severe" (20+). *Anxiety symptoms* were assessed using the self-report Generalised Anxiety Disorder Scale (GAD-7) (33). This seven-item questionnaire assessed the presence of generalised anxiety symptoms in the past two weeks. It uses the same response scale as the PHQ-9 and participants' total scores can also be classified into the same severity categories. Participants were also asked if they had had a panic attack in the past two weeks, and if so, how many. *Suicidal thoughts scores* were based on participants' responses to item 9 of the PHQ-9 (ranges from 0 to 3). Participants who reported that they experienced "thoughts that they would be better off dead, of harming themselves" for more than several days (score > 0) were deemed to have suicidal thoughts.

**Social media data extraction and analysis.** Social media data was extracted every fortnight using the publicly accessible Application Programming Interface (APIs) for each platform, including Tumblr, Live Journal, WordPress and BlogSpot. Social media data was analysed using the Linguistic Inquiry and Word Count (LIWC) tool for linguistic features. This software analyses text and calculates the percentage of words that reflect different emotions, thinking styles, social concerns, and parts of speech (34), resulting in a set of 68 linguistic features for each blog post.

**Data analysis.** We first performed bivariate analyses to investigate the correlation between the linguistic features and mental health scores between individuals. To this end, we averaged linguistic features and mental health scores across repeated assessments for each participant. We first performed bivariate analysis between the 68 linguistic features and the 3 mental health scores using Spearman's rank-order correlation. Hence, a total of $68 \times 3 = 204$ comparisons were performed. We used permutation tests to control the family-wise error rate (35, 36). A permutation was constructed by exchanging the mental health scores across participants and a new correlation coefficient was recomputed for the permuted data. This process was repeated for 10,000 permutations, resulting in the distribution of possible correlations coefficients for these data under the null hypothesis that observations are exchangeable. This procedure can be generalised to a family of similar tests by computing the distribution of the most extreme statistic (here the most extreme positive or negative correlations coefficient) across the entire family of tests for each permutation. This procedure corrects for multiple comparisons because the distribution of extreme statistics, from which the p-values of each comparison are derived, automatically adjusts to reflect the increased chance of false discoveries due to an increased number of comparisons (36). We used the Matlab function *mult_comp_perm_corr.m* (https://au.mathworks.com/matlabcentral/fileexchange/34920-mult-comp-perm-corr) to perform the mass bivariate analyses. We used bootstrapping to estimate the confidence interval for the correlation coefficients (37).

**Multivariate regression.** We then performed multivariate analysis between multiple linguistic features using partial-least squares (PLS) regression. PLS regression is a multivariate extension of linear regression that builds prediction functions based on components extracted from the co-variance structure between features and targets (38). Although it is possible to calculate as many PLS components as the rank of the target matrix, not all of them are normally used as data are never noise-free and extracting too many components will result in overfitting. Both the features and targets were z-transformed before performing PLS regression analysis. We used 5-fold cross-validation (39) to determine the number of components of the model and prediction accuracy was assessed using Mean Square Error (MSE). To select the best model, the number of components was increased until the MSE not further decreased. We used the build-in Matlab function *plsregress.m* from the Statistics and Machine Learning Toolbox (R2018a) to perform PLS regression. We performed PLS regression both on the full and a restricted feature set. To obtain a restricted feature set, we used bootstrapping: The z-scores of the feature loadings were estimated across 10,000 bootstrap samples and the four features with the largest z-score were selected. The PLS regression procedure was then repeated using only these four features.

**Within-subject prediction.** The previous analyses constructed regression models that predicted the mental health scores of participants that were not included in the training set. These group-level inferences do not necessarily generalise to intra-individual changes in mental health scores over time (28). We therefore tested the PLS regression model constructed on group-level data on repeated measures of single participants using a two-staged approach. We first used the group-level model to predict the mental scores at each time point at which linguistic features were extracted and mental health were assessed and correlated the predicted and observed mental health scores across time points for each participant. We then compared the correlation coefficients estimated for each participant at the group level. To do this, we converted the correlation coefficients using Fisher's z transformation and compared the z-scores against zero using a one-sample t-test.

**Availability of data and code**. The data used in this study is available at Zenodo: https://doi.org/10.5281/zenodo.1476493. The data includes the linguistic features extracted from the blog post and the mental health scores of each assessment. The Matlab scripts used to perform the bivariate and multivariate analyses are available at Zenodo: https://doi.org/10.5281/zenodo.1476505.

**RESULTS**

**Participants**

A total of 153 individuals consented to the study and completed the baseline assessment (88% female, mean age: 29.5 years, SD: 10.3, age range: 18-67). The mean number of completed assessments was 5.1 (SD: 4.4). See the Supplementary Material for more information about these participants. The final sample consisted of the 38 participants who completed one or more mental health assessments and who also blogged at least once during the study period. The total number of blog posts collected from the final sample was 655. Table 1 outlines participant characteristics at baseline.

**Table 1. Participant characteristics at baseline (N=38)**

|  | M (SD) | Range |
|---|---|---|
| Age | 27.1 (7.2) | 18-59 |
| PHQ-9 score | 14.5 (5.6) | 3-27 |
| GAD-7 score | 11.6 (5.0) | 0-20 |
| Suicide/Self-harm score | 1.07 (1.1) | 0-3 |
| Panic attacks | 7.0 (11.1) | 0-60 |
| Number of blog posts | 17.2 (26.7) | 1-128 |
|  | N | % |
| Female | 32 | 84 |
| Self-rated health |  |  |
|   Very bad | 0 | 0 |
|   Bad | 11 | 29 |
|   Moderate | 15 | 39 |
|   Good | 12 | 32 |
|   Very good | 0 | 0 |
| Prior diagnosis by medical practitioner | 36 | 95 |
| Currently taking medication | 26 | 68 |
| Current suicidal thoughts | 25 | 66 |
| Panic attack in 2 weeks prior | 22 | 58 |
| Depression |  |  |
|   Nil-minimal | 2 | 5 |
|   Mild | 6 | 16 |
|   Moderate | 10 | 26 |
|   Moderately-severe | 11 | 29 |
|   Severe | 9 | 24 |
| Anxiety |  |  |
|   Nil-minimal | 3 | 8 |
|   Mild | 9 | 24 |
| Moderate | 15 | 39 |
| Moderately-severe | 9 | 24 |
| Severe | 2 | 5 |
| Blog site used | | |
| LiveJournal | 30 | 79 |
| Tumblr | 8 | 21 |

On average, participants had moderately severe levels of depression and anxiety, but symptoms varied considerably between participants (PHQ-9, SD: 5.7, range: 1.8-26.0; GAD-7, SD: 4.8, range: 0-20.7; Fig. 1A). Intra-individual differences between mental health scores showed a much smaller spread (Fig. S1): the mean of the intra-individual standard deviation was 2.8 (SD: 2.7, range: 0-10.1) for the PHQ-9 and 2.2 (SD: 2.0, range: 0-7.6) for the GAD-7. Figure 1B shows the number of mental health assessments completed across the study period, with only 2 completing all 18.

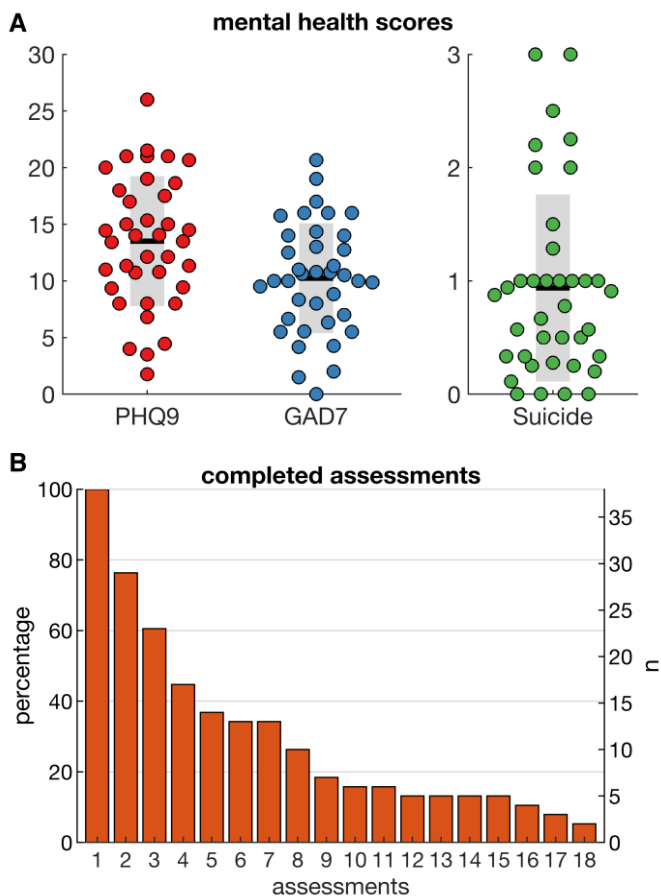
11

**Figure 1. Characteristics of sample A)** Mental health scores (PHQ-9, GAD-7 and suicidal ideation) averaged across assessments. Coloured dots show values of individual participants, the horizontal black line the group mean and the grey bars the SD. **B)** Number of mental health assessments completed by participants.

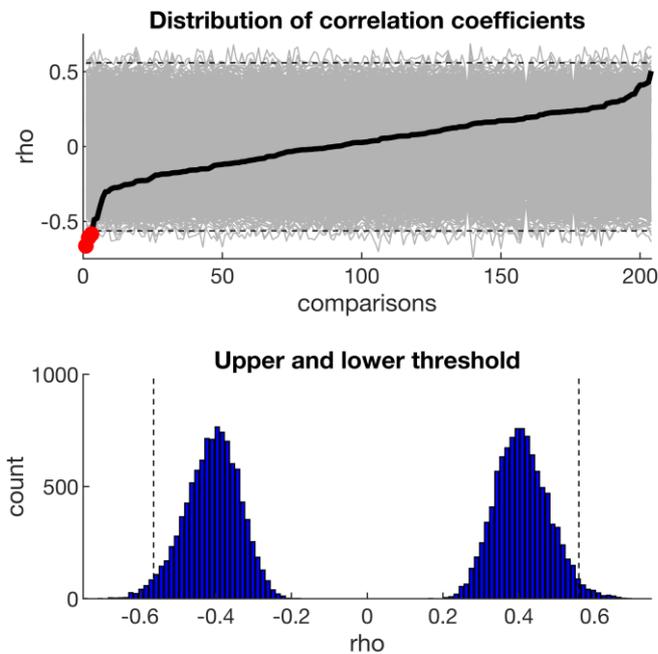

**Figure 2. Permutation testing to control the family-wise error rate of mass bivariate analyses.** Top panel shows the correlation coefficients for all 204 comparisons (black line) as well as the correlation coefficients of the 10,000 permutations (grey lines). The distribution of the most extreme statistic (bottom panel) was then used to determine adjusted significance threshold (dash line). Using this threshold only a few correlations are statistically significant (red dots, top panel).

**Between-subjects analysis**

We first performed mass bivariate analyses correlating 68 linguistic features with the 3 mental health scores. We used permutation tests to compute the distribution of the most extreme statistic across all comparisons and control the family-wise error rate. The resulting significance threshold was $|rho| > 0.56$ (Fig. 2). Tables 2 outlines the linguistic features which showed the strongest correlations with depression, anxiety, and suicidal thoughts,

respectively. After controlling for multiple comparisons, only non-fluencies was significantly correlated with depression (rho = -0.61, 95%CI: -0.77, -0.39, $P_{corr}$ = 0.012), and non-fluencies and tentativeness with anxiety (rho = -0.58, 95%CI: -0.75, -0.31, $P_{corr}$ = 0.028 and rho = -0.67, 95%CI: -0.79, -0.47, $P_{corr}$ = 0.002, respectively).

**Table 2. Between-subject analysis**

| target | feature | example | rho | 95% CI | $P_{uncorr}$ | $P_{corr}$ |
|---|---|---|---|---|---|---|
| depression | present focus | *is, does, here* | 0.42 | 0.14, 0.62 | 0.008 | 0.84 |
| | sadness | *crying, grief, sad* | 0.41 | 0.12, 0.63 | 0.010 | 0.93 |
| | certainty | *always, never* | 0.39 | 0.08, 0.63 | 0.016 | 1 |
| | 3rd person plural | *they, their, they'd* | -0.34 | -0.59, -0.04 | 0.034 | 1 |
| | tentative | *maybe, perhaps,* | -0.49 | -0.66, -0.23 | 0.002 | 0.28 |
| | non-fluencies | *er, hm, umm* | -0.61 | -0.77, -0.39 | 0.000 | **0.012** |
| anxiety | present focus | *is, does, here* | 0.50 | 0.21, 0.71 | 0.001 | 0.21 |
| | 1st person singular | *I, me, mine* | 0.35 | 0.06, 0.60 | 0.031 | 1 |
| | total pronouns | *I, them, itself* | 0.30 | 0, 0.56 | 0.068 | 1 |
| | future focus | *will, gonna* | -0.41 | -0.66, -0.08 | 0.011 | 0.96 |
| | tentative | *maybe, perhaps* | -0.58 | -0.75, -0.31 | 0.000 | **0.028** |
| | non-fluencies | *er, hm, umm* | -0.67 | -0.79, -0.47 | 0.000 | **0.002** |
| suicidality | ingesting | *dish, eat, pizza* | 0.43 | 0.13, 0.66 | 0.007 | 0.74 |
| | impersonal pronouns | *it, it's, those* | 0.41 | 0.09, 0.69 | 0.010 | 0.94 |
| | certainty | *always, never* | 0.32 | 0, 0.58 | 0.047 | 1 |
| | relativity | *area, exist, stop* | -0.28 | -0.53, 0.02 | 0.093 | 1 |
| | space | *down, in, thin* | -0.28 | -0.54, 0.01 | 0.084 | 1 |
| | non-fluencies | *er, hm, umm* | -0.48 | -0.69, -0.20 | 0.002 | 0.32 |

We then performed PLS regression to extract multiple linguistic features that predict mental health scores. We first used all 68 linguistic features and used 5-fold cross-validation to determine the optimal number of components. For PHQ-9 and GAD-7, a PLS model with one component revealed a reduction in MSE (-9% and -2%, respectively), while no PLS model showed a reduction in MSE compared to a model with zero components for suicidal thought (Fig. 3, left column). We used bootstrapping to determine the features that were most robust across participants. The four features with the highest absolute z-scores were *3$^{rd}$ person pronouns, present focus, quantifiers (e.g. few many, much)* and *tentative (e.g. maybe, perhaps)* for PHQ-9, *present focus, 1$^{st}$ person singular, tentative* and *3$^{rd}$ person plural* for GAD-7, and *dictionary words, axillary verbs (e.g. am, will, have), 1$^{st}$ person singular pronouns,* and *negations (e.g. no, not, never)* for suicidal ideation. Most of these features were among the features with the strongest correlations in the bivariate analyses (Table 2).

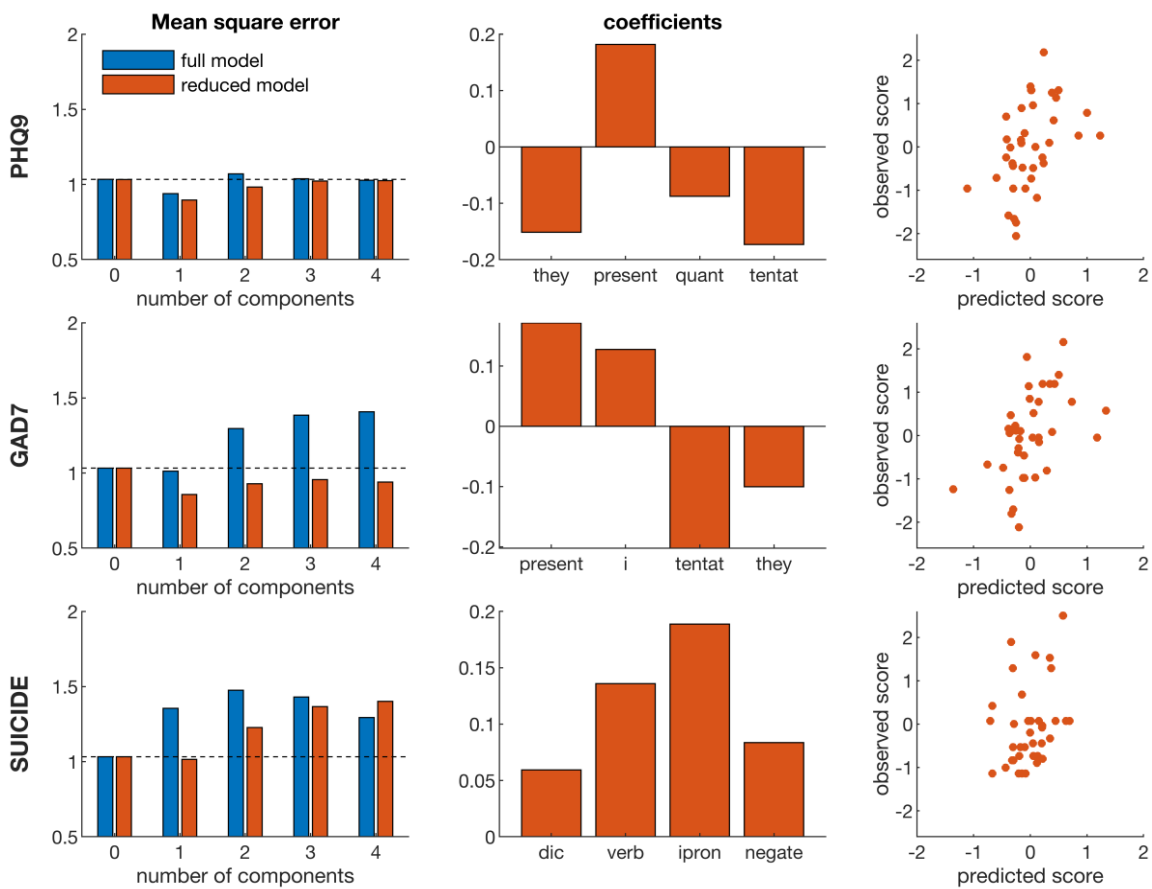

**Figure 3. Results of PLS regression at group level.** We used 5-fold cross-validation to determine the number of PLS components. The optimal model was the model showing the lowest MSE (left panel).

We tested both the full model using all 68 linguistic features and a reduced model using only the 4 most robust features. The beta coefficients of the optimal model (middle column) were then used to estimate the predicted mental health scores (right column). Note: they=3$^{rd}$ person plural, present=present focus, quant=quantifiers, tentat=tentative, i=1$^{st}$ person pronouns, dic=dictionary words, verb=auxiallary verbs, ipron= impersonal pronouns, and negate=negative emotion.

We then tested the reduced PLS models using only the four most robust features. The reduced models showed a larger reduction in MSE during cross-validation than the full PLS model: -13% for PHQ-9, -17% for GAD-7, and -1% for suicidal ideation (Fig. 3, left column). We used the reduced PLS model with one component, as the MSE increased again when adding additional components. Figure 3 shows the beta coefficients of the regression models (middle column) and the predicted mental health scores (right column). The correlation between the predicted and observed mental scores is r = 0.44 (95%CI: 0.14, 0.67, $R^2$ = 0.20) for PHQ-9, r = 0.49 (95%CI: 0.20, 0.70, $R^2$ = 0.24) for GAD-7, and r = 0.36 (95%CI: 0.04, 0.61, $R^2$ = 0.13) for suicidal ideation. A PLS model combining the three targets revealed that the mental health scores are correlated and can be predicted using the same linguistic features (Fig. S2), although the reduction in RMS (-7%) is smaller than for the models predicting individual mental health scores.

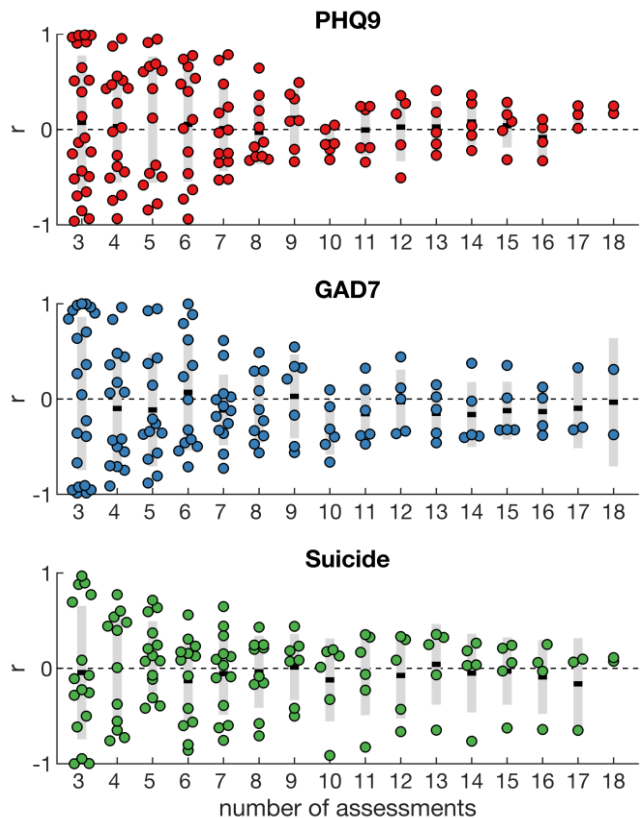

**Figure 4. Intra-individual correlations.** Group-level regression models were used to predict intra-individual changes in mental health scores. The predicted mental health scores were correlated with the observed mental health scores for each participant. The coloured dots show the correlation coefficients for individual participants, the black line the group mean and the grey bars the 95% CI. The distribution of correlation coefficients was estimated for all participants having completed at least 3 to 18 assessments. The number of participants decreased with increasing number of assessments, as participants did not complete all assessments (see Fig. 1B).

**Within-subjects analysis**

We used the regression models estimated from group-level data to predict within-subject variations in mental health scores over time. As participants completed different numbers of assessments (see Fig. 1B), we tested the models multiple times for all participants having completed at least *n* assessments. The distribution of correlations coefficients across participants fluctuated around zero (Fig. 4). When the correlation coefficient was estimated over a larger number of assessments the distribution become narrower; however, the 95% CI generally overlapped with zero. As such, the positive correlations observed at group level (Fig. 3) were not observed for within-subject correlations. In fact, the average correlation coefficient for GAD-7 was negative when estimated over at least 10 repeated assessments.

**DISCUSSION**

This study investigated the relationship between linguistic features in blog content and individuals' mental health scores for depression, anxiety, and suicidal thinking, over a 32-week period. This study examined both group-level and individual-level correlations to test whether linguistic expression in blogs can be used to determine the mental state of individuals who use these platforms. We found mixed evidence for the hypotheses.

In the bivariate analyses of between-subjects correlations, only two linguistic features emerged as significant when controlling the family-wise error rate: tentativeness and non-fluencies. Tentativeness, which is the degree of uncertainty reflected in text, was associated with anxiety only. This may reflect the increased worry and hesitation experienced by those who are anxious. Non-fluencies were associated with depression and anxiety symptoms, but not suicidal thoughts. In speech, non-fluencies relate to the various breaks and irregularities in composition and reflects pauses for thought, nervousness, or decreased alertness (40). These have been found to be greater in depressed people (41, 42) and may reflect the cognitive deficits associated with the illness. However, little is known about depression and the fluency of written text. In this study, participants with higher depressive and anxious symptoms showed fewer non-fluencies in their written blog content. This finding and its direction is not consistent with past studies on speech patterns or social media expression. It may represent initial evidence of modality specific features (18), as blogs allow individuals to write longer sections of text. Further examination in other datasets would confirm this.

Multivariate modelling revealed some similarities with the bivariate results, but also key differences as quantifiers, dictionary words, axillary verbs, and negations emerged as robust predictors of mental health. The differences with the bivariate analyses are partly explained by the type of correlation on which these analyses are based: we used rank-order correlation for the bivariate analyses while PLS regression finds a linear regression model. Some of the features, such as non-fluencies, were not normally distributed and hence showed significant rank-order correlation but were not among the robust features of the linear PLS model. Here we used basic regression techniques to assess the relationship between linguistic features and mental health states. More advanced machine learning techniques may be used to improve the prediction of mental health state from automated text analyses (43), but more complex models generally require large datasets to avoid overfitting (44, 45). Indeed, in the current study we found that a PLS model based on a single component had a lower prediction error than models with additional components (Fig. 3).

The current findings are somewhat consistent with the previous studies that have used validated psycho-metric scales of depression alongside social media data. The features of third person pronouns, present focus, 1$^{st}$ person singular are consistent with De Choudhury et al (13), providing further support as a marker of the distancing from others and focus on oneself that occurs in a depressive and suicidal state. The increased usage of first person singular pronouns as a marker of depression is also consistent with Eichstaedt et al (19) and Edwards & Holtzman's meta-analysis (21). This suggests that the markers of depression found in traditional forms of text such as poetry and letters are also likely to be evident in online communication. In contrast to Tsugawa et al (15), negative emotion did not emerge as significant for depression, anxiety, or suicidal ideation. There was also little overlap with the features (e.g. word count, ingestion, sadness) found by Reece et al (14). However, it is difficult to compare our findings with past studies due to differences in modelling as not all features were measured, and varying analytical techniques were used. Further, the duration of data collection varies across studies. This emerging area of research will benefit significantly from replication studies in which the same features and models are used across various datasets.

An important strength of this study was the longitudinal design. This allowed us to examine the relationship between mental health symptoms and linguistic expression over time. We expected that individual symptoms would fluctuate over the 32-weeks and would be associated with a change in linguistic expression. However, this was not the case. The correlations identified at the group-level were not significant at the individual level. Thus, our findings do not support group-to-individual generalisability of linguistic markers of depression, anxiety, and suicidal thinking. Indeed, previous research has shown that relationships observed at the group level do not necessarily generalise to all the individuals in this sample and hence requires explicit testing (28). In part, the lack of significant within-subjects correlations in our study may be due to missing data as not all participants completed all 18 assessments, which reduced the statistical power of the analyses we performed. Moreover, some participants show little variation in mental health scores over time (Fig. S1). However, the lack of group-to-individual generalisability may also indicate that the underlying processes are non-ergodic (28), that is, the relationship between linguistic features and mental health state may not be equivalent across individuals and time. The relationship between linguistic features and mental health state may be specific to subgroups, such as the type of mental health concern, or demographics such as gender and age, or the type of blog post. We should hence carefully examine, rather than assume, whether relationships observed at the group level also hold for subgroups or individuals over time (30). Hence, previous findings showing correlations between linguistic features and mental health states at group level cannot be assumed to generalise to individuals. Intensive repeated-measures data in larger samples are needed to make predictions about intra-individual changes in mental health scores over time.

**Limitations**

While the design used in this study has the potential to inform knowledge on mental health symptoms and the relationship to linguistic expression in social media across time, it was hampered by low levels of data. There was significant drop-out, non-completion of mental health assessments, and variability in the amount of blog data that was generated by participants. While attrition is common and seemingly unavoidable in longitudinal studies (46), the identification of markers of mental ill-health requires large amounts of individual data collected over long periods of time. This burden may be alleviated by sharing data and results, the demand for which has been increased by open access (47). As it can be challenging to engage human research subjects in sustained studies, and the effects of repeated measures in psychiatry is still unknown, sharing data between researchers could alleviate these burdens. Moreover, open access provides an opportunity to test models on datasets from multiple sources and platforms and test whether predictions generalise to new data (39). The high number of features has the danger of inflating researcher degrees of freedom and may endanger replicability of findings (48). Practices such as preregistration of study hypotheses and methods could help reduce spurious correlations and will be key in identifying reliable markers of mental health state (49). Testing predefined models to new data is likely to be the primary way for the field to advance. We therefore shared the data of the linguistic features and mental health scores that we acquired to inform further studies or provide independent testing data for existing prediction models.

## CONCLUSIONS

Social media presents an exciting opportunity for developing new tools to monitor the onset of mental illness. This study examined the associations between linguistic features in blog content and individuals' self-reported depression, anxiety, and suicidal ideation. Several features were significantly associated with mental health scores when assessed across participants, with differences found between the bivariate and multivariate analyses. Cross-validation showed that linguistic features can predict the mental health scores of participants that were not included in the training set. When testing the multivariate regression models on longitudinal data of individual participants, no robust correlations were found between changes in linguistic features and mental health scores over time. These finding shows that linguistic features can identify individuals with mental illness but may not be able to detect individual changes in mental health over time. This study demonstrates the importance of a longitudinal study design and the use of validated psychometric scales. Future studies, utilising the advantages of open access, will need to confirm whether social media data can also be used to predict individual changes in mental health over time.

## ACKNOWLEDGEMENTS

HC and this research were financially supported by NHMRC John Cade Fellowship APP1056964. BOD and MEL were supported by the Society for Mental Health Research (SMHR) Early Career Researcher Awards. TB was supported by a NARSAD Young Investigator Grant from the Brain & Behavior Research Foundation.

**SUPPLEMENTARY MATERIAL**

**Background information on all consenting participants:**

Of the 153 participants who consented to take part in the study, 127 (83%) had received a diagnosis of depression or anxiety from a medical practitioner, and 87 (57%) were taking medication for depression or anxiety. For depression scores, 22 were nil-minimal (14%), 36 mild (24%), 29 moderate (19%), 39 moderately severe (25%) and 27 severe (18%). For anxiety scores, 34 were minimal (22%), 47 mild (31%), 40 moderate (26%), 25 moderately severe (16%) and 7 severe (5%). A total of 72 (47%) reporting having had an anxiety attack in the two weeks prior, with the mean number of attacks in this period being 5.3 (SD:7.4, range: 1-60). At baseline, 81 (53%) had "thoughts that they would be better off dead, or of hurting themselves" for several days or more in the past two weeks. Also, at baseline, 68 (44%) reported both moderate to severe depression and anxiety, with depression and anxiety scores highly correlated ($r = 0.78$, $p < .001$).

**Intra-individual variability**

The current study collected mental health data longitudinally and participants were asked to complete the PHQ-9 and GAD7 every two weeks for a total of 18 assessments. To assess whether participants revealed longitudinal changes in their mental health levels, we assessed variability between repeated assessments. Participants completed 5.4 assessments on average (SD: 5.2, Fig. 1B). The patterns of longitudinal change differed across participants. Several participants revealed large changes in mental health scores over time: 12 out of 38 participants had at least a 10-point differences between the lowest and highest PHQ-9 score and 7 out of 38 participants had at least a 10-point differences between the lowest and highest GAD-7 score (Fig. S1). As the Suicidal thoughts scores were based on participants' responses to item 9 of the PHQ-9 which ranges from 0 to 3, the longitudinal change for this target was lower.

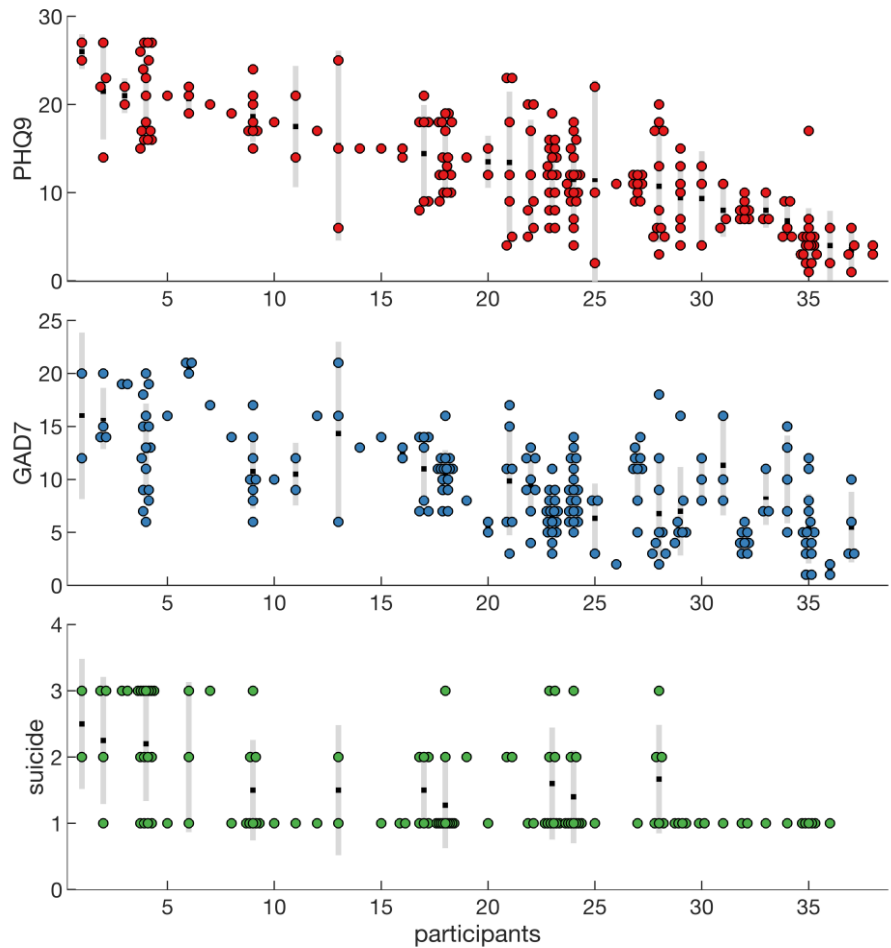

**Figure S1**. Repeated mental health scores for individual participants (n=38). Coloured dots show values of repeated assessments, the horizontal black line the individual mean and the grey bars the SD.

**Single PLS model to predicted combined mental health outcomes**

We assessed PLS regression models for each mental health target (PHQ-9, GAD-7 and suicide) separately. However, PLS regression also allows to predict multiple target measures using a single model. We also assessed the PLS model that predicted all three mental health scores simultaneously. Similar to the separate PLS models, the combined model showed the lowest MSE for a model with a single component. With a single component, the full model showed an increase in MSE of 0.08 (+2.6%), while the reduced model showed a reduction in MSE of 0.23 (-7.3%; Fig. S2). The reduced model explained 15% of the variance. The positive weights for the three targets reflect the positive correlations between the three mental health scores.

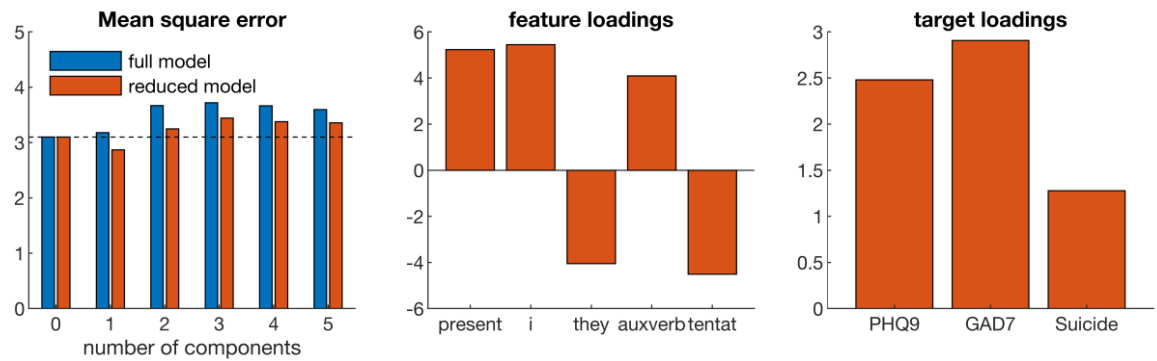

**Figure S2**. Results of PLS regression for combined mental health outcomes. We used 5-fold cross-validation to determine the number of PLS components. The optimal model was the model showing the lowest MSE (left panel). We tested both the full model using all 68 linguistic features and a reduced model using only the 5 most robust features. The middle panel shows the beta coefficients of the most robust features and the right column the coefficients of the three targets. Note: present=present focus, i=1st person pronouns, they=3rd person plural, auxverb=auxiliary verbs, and tentat=tentative.